\documentclass{l4dc2024}

\usepackage{graphics} 
\usepackage{mathptmx} 
\usepackage{times} 
\usepackage{enumitem}
\usepackage{standalone}
\usepackage[utf8]{inputenc}
\usepackage{pgfplots}
\DeclareUnicodeCharacter{2212}{−}
\usepgfplotslibrary{groupplots,dateplot}
\usetikzlibrary{patterns,shapes.arrows}
\pgfplotsset{compat=newest}

\usepackage{optidef}
\usepackage{comment}

\newenvironment{eqs*}{\begin{equation*}\aligned}{\endaligned\end{equation*}}

\DeclareMathOperator*{\minimize}{minimize}				
\newcommand{\mat}[1]{\ensuremath{\begin{bmatrix}#1\end{bmatrix}}}	

\newcommand{\Rv}[1]{\ensuremath{\mathbb{R}^{#1}}}				
\newcommand{\R}[2]{\ensuremath{\mathbb{R}^{#1\times #2}}}		
\newcommand{\Expect}{{\rm I\kern-.3em E}}				

\usepackage{xcolor}

\newcommand{\ea}[1]{\textcolor{orange}{\bf [EA: #1]}}

\title[CACTO-SL]{CACTO-SL: Using Sobolev Learning to improve\\Continuous Actor-Critic with Trajectory Optimization}

\author{%
\Name{Elisa Alboni$^{1}$}
\Email{elisa.alboni@unitn.it}
\AND
\Name{Gianluigi Grandesso$^{1}$}
\Email{gianluigi.grandesso@unitn.it}
\AND
\Name{Gastone P. {Rosati Papini}$^{1}$}
\Email{gastone.rosatipapini@unitn.it}
\AND
\Name{Justin Carpentier$^{2}$}
\Email{justin.carpentier@inria.fr}
\AND
\Name{Andrea {Del Prete}$^{1}$}
\Email{andrea.delprete@unitn.it} \\
\addr $^{1}$ Dept. of Industrial Engineering, University of Trento, Italy \\
\addr $^{2}$ INRIA, Paris, France%
}

\begin{document}

\maketitle

\begin{abstract}%
Trajectory Optimization (TO) and Reinforcement Learning (RL) are powerful and complementary tools to solve optimal control problems. 
On the one hand, TO can efficiently compute locally-optimal solutions, but it tends to get stuck in local minima if the problem is not convex. 
On the other hand, RL is typically less sensitive to non-convexity, but it requires a much higher computational effort. 
Recently, we have proposed CACTO (Continuous Actor-Critic with Trajectory Optimization), an algorithm that uses TO to guide the exploration of an actor-critic RL algorithm.
In turns, the policy encoded by the actor is used to warm-start TO, closing the loop between TO and RL. 
In this work, we present CACTO-SL, an extension of CACTO exploiting the idea of Sobolev Learning. 
To make the training of the critic network faster and more data efficient, we enrich it with the gradient of the Value function, computed via a backward pass of the differential dynamic programming algorithm.
Our results show that the new algorithm is more efficient than the original CACTO, reducing the number of TO episodes by a factor ranging from 3 to 10, and consequently the computation time. Moreover, we show that CACTO-SL helps TO to find better minima and to produce more consistent results.%
\end{abstract}


\section{Introduction}
Robot control challenges have long been addressed through Trajectory Optimization (TO). The high-level desired task is encoded in the cost function of an Optimal Control Problem (OCP), which is minimised with respect to the OCP decision variables, which represent the state and control trajectories. Constraints are added to ensure that the solution of the OCP takes into account the robot dynamics, the actuator bounds, and the task-related constraints. When tackling complex problems, the OCP may feature a highly non-convex cost and/or highly nonlinear dynamics. Therefore, gradient-based solvers frequently encounter local minima and are unable to find a globally optimal solution. There exist TO methods based on the Hamilton-Jacobi-Bellman equation or Dynamic Programming, for continuous-time and discrete-time problems, respectively, that can find a globally optimal solution. However, these methods are hindered by the curse of dimensionality, which limits their applicability. 

With the emergence of deep Reinforcement Learning (RL) and its application to the continuous domain, this machine learning tool is applied more and more widely to robot control problems showing impressive results on continuous state and control spaces~\cite{ddpg,td3,sac,ppo}. RL algorithms are less prone to converge to local minima due to their exploratory nature. Yet, there are still several challenges related to the application of RL to robot control, such as the necessity for extensive exploration.

As a potential solution to overcome the limitations of RL and TO, we have recently presented the CACTO algorithm (Continuous Actor-Critic with Trajectory Optimization)~\cite{CACTO}. CACTO iteratively leverages the explorative nature of RL to initialize TO to escape local minima, while exploiting TO to guide the RL exploration. Thanks to the interplay of TO and RL, CACTO's policy provides TO with initial guesses that allow it to obtain better trajectories than with other initialization techniques. 
While the use of TO demonstrated to efficiently accelerate convergence of RL, the computational burden associated with solving TO episodes has posed some limitations. In particular, as the system complexity increases, this issue becomes more relevant, hindering the scalability of the algorithm. The ability of TO to produce the Value function's gradient with limited computational cost presents an opportunity to enhance the algorithm's data efficiency, by increasing the information extracted from each TO problem. By leveraging Sobolev Learning (SL), the gradient information can be incorporated in the critic's training, enhancing the performance of the algorithm.

Sobolev spaces are metric spaces where the distance between functions is defined in terms of both the difference between the function values and the difference between their derivatives values. The universal approximation theorem for neural networks in Sobolev spaces~\cite{Hornik} shows that, under some assumptions, not only a neural network can approximate the value of a function, but also its derivatives with respect to its inputs. This work inspired other research, such as \cite{SobolevDM}, which extensively studied the employment of the neural network derivatives to improve the training process in different kinds of problems, including policy distillation and regression on datasets. Sobolev Learning found applications also in the robot control field. For example, in \cite{SobolevLaas}, this technique is used to learn a Value function to be used as an approximate terminal cost in an OCP, thus allowing to shorten the problem horizon, and speeding up the solver. In \cite{SobolevLaas2}, stochastic Sobolev Learning is used to include computationally efficient higher-order information in the policy training, resulting in improved sample efficiency and stability. Even though the use of the derivatives comes with a computational overhead, Sobolev Learning increases robustness against noise and improves generalization as well as data-efficiency \cite{noise,generalization}. In this work, we incorporate Sobolev Learning in CACTO to improve its efficiency and scalability. 
Our main contributions are:
\begin{itemize}
    \item We present a new version of the CACTO algorithm that computes the gradient of the Value function using the backward pass of the differential dynamic programming algorithm, and uses it to improve the training of the critic network. 
    \item We show that using ReLU activation functions is detrimental when training the critic with Sobolev Learning, and we address this issue by switching to smooth periodic activation functions.
    \item We integrated our open-source implementation of CACTO with the software libraries \emph{CasADi}~\cite{casadi} (for numerical optimization) and \emph{Pinocchio}~\cite{Pinocchio} (for multi-body dynamics), making it more versatile and easily accessible by the community.
    \item We exemplify the behavior of our algorithm in the companion video using a simple 1D toy problem.
\end{itemize}

\section{Method}
In \cite{CACTO}, we presented CACTO, an algorithm for finding globally-optimal control policies through TO-guided actor-critic RL, which we summarize in Section~\ref{ssec:cacto}. 
To increase its computational efficiency, we propose to use Sobolev Learning~\cite{SobolevDM} for training the critic network, as detailed in Section~\ref{ssec:cacto-sl}. 


\subsection{Original CACTO Formulation}
\label{ssec:cacto}
This section presents the former formulation of CACTO, an optimization algorithm designed to address discrete-time optimal control problems with a finite time horizon, with the following structure: 
\begin{mini!}|l|[2]<b>
{\scriptstyle{X, U}}
{L(X,U) = \sum_{t=0}^{T -1} l_t\left(x_t,u_t\right) + \ l_T\left(x_T\right) 
\label{OCP_objective}}
{\label{eq:ocp}}
{}
\addConstraint{x_{t+1}}{= f(x_t,u_t)  \quad \; \forall t=0\dots T-1\label{OCP_dyn_const}}
\addConstraint{|u_t|}{\le u_{max} \quad\quad\quad \, \forall t=0\dots T-1\label{OCP_path_const}}
\addConstraint{x_0}{= x_{init}\label{OCP_ICS}}
\end{mini!}
where the decision variables are the state and control sequences denoted as $X=x_{0\dots T}$ and $U=u_{0\dots T-1}$, with $x_t \in \mathbb{R}^{n}$ and $u_t \in \mathbb{R}^{m}$. The cost function $L(\cdot)$ is defined as the sum of the running costs $l_t\left(x_t,u_t\right)$ and the terminal cost $l_T(x_T)$. The dynamics, control limits and initial conditions are represented by \eqref{OCP_dyn_const}, \eqref{OCP_path_const} and \eqref{OCP_ICS}.

The algorithm begins with the TO-phase, where it solves $N$ TO problems~\eqref{eq:ocp} with random initial states $x_{init}$, random time horizons in $[1, T]$, and using a classic warm-starting technique, e.g. initializing states to $x_{init}$ and control inputs to zero. 
For each optimal state computed by TO, we compute the partial $L$-step cost-to-go, where $L$ is the number of lookahead steps used for Temporal Difference (TD) learning, and store them in a replay buffer along with the relative transition (i.e., state, control, and state after $L$ steps). The states stored in the replay buffer are augmented states $\Tilde x = [x,t]$, where $t$ is the time.
Next we start the update phase: for $M$ times, a batch of $N_B$ transitions is sampled from the replay buffer and used to update the neural networks of the critic and the actor. 
In particular, the critic loss is the mean squared error between the critic's output and the so-called TD target, while the actor loss is the action-value (i.e., Q) function. 
Finally, a rollout of the actor's policy is used to warm-start the TO problems in the next episodes, closing the loop.
In \cite{CACTO}, we showed the capabilities of CACTO to escape local minima, while being more computationally efficient than state-of-the-art RL algorithms. Moreover, CACTO has been proven to converge to a global optimum in a discrete-space setting.

\subsection{CACTO with Sobolev Learning (CACTO-SL)} \label{ssec:cacto-sl}
\begin{figure}[tbp]
 \centering
 \includegraphics[width=0.75\textwidth]{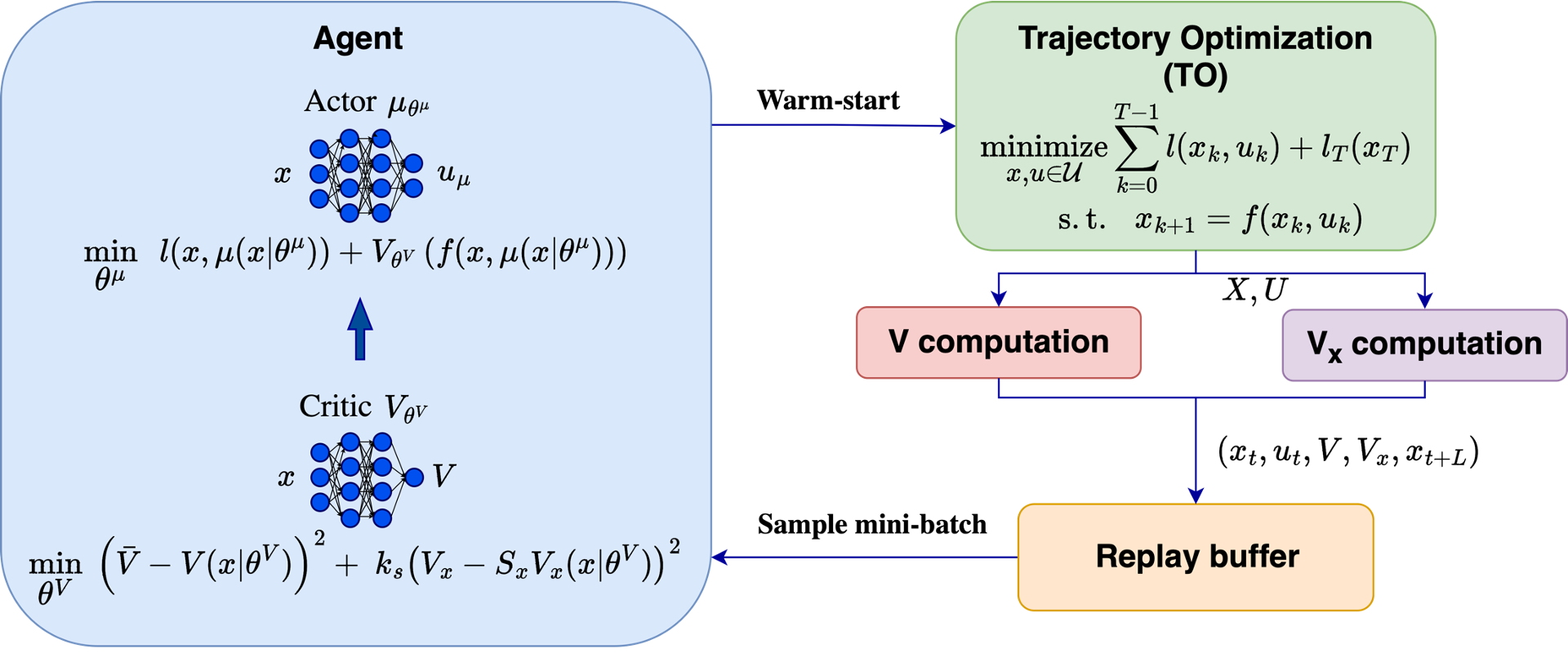}
 \caption{Scheme of the CACTO-SL algorithm.}
 \label{fig:CACTO-SL_scheme}
\end{figure}
To exploit Sobolev Learning we need to provide the gradient of the Value function with respect to the state: $V_x$. To analytically compute $V_x$, we use the backward-pass of \textit{Differential Dynamic Programming} (DDP)~\cite{jacobson1968new, Tassa2012}, an efficient optimal control method for unconstrained nonlinear problems. 
A scheme of the CACTO-SL algorithm is depicted in Fig.~\ref{fig:CACTO-SL_scheme}.
Notice that, since DDP does not handle control bounds, the OCPs solved with CACTO-SL do not feature control bounds, but penalties.

\subsubsection{Using the gradient of the Value function in CACTO-SL}
To exploit $V_x$ when training the critic, we store it in the replay buffer along with the associated transition. 
Then, during the update-phase, the critic parameters $\theta^V$ are updated to match both the target values $\bar{V}$ and their gradients $V_x$:
\begin{equation}
\minimize_{\theta^V} \quad 
\frac{1}{N_B}\sum\limits_{i=1}^{N_B} \left(\bar{V}_i - V(\Tilde x_i|\theta^V) \right)^2 +
k_S \left(V_{x,i} - S_{x} V_{\Tilde x}(\Tilde x_i|\theta^V) \right)^2,
\label{eq:critic_loss}
\end{equation}
where $N_B$ is the batch size and the matrix $S_{x} = \mat{I_{n\times n} & 0_n} \in \R{n}{n+1}$ selects the first $n$ elements of the vector $V_{\Tilde x}(\Tilde x_i|\theta^V) \in \Rv{n+1}$, thus excluding from the error the derivative of $V$ w.r.t. $t$.
This is because the state in CACTO is augmented with the time $t$, therefore the gradient of the critic contains also the derivative w.r.t. $t$. 
However, DDP does not compute it because it uses a discrete representation for the time.

The relative weight of the two components in~\eqref{eq:critic_loss} is defined by the coefficient $k_S$. 
Tuning $k_S$ to achieve the right trade-off is important for optimal performance. 
As $k_S$ gets smaller, CACTO-SL tends to the original CACTO algorithm. 
Setting $k_S$ to a large value may seem sensible because the training of the actor, which is our ultimate goal, relies solely on $V_x$. 
However, we should consider that the Value function is likely discontinuous at the boundaries of the basins of attraction of the different locally-optimal solutions of the OCP.
Around discontinuities, the gradient fails to capture the local behavior of the function, therefore setting $k_S$ too large may result in extremely poor approximations of the Value around discontinuities.
This could slow down, or even prevent, convergence to a globally-optimal policy. 

To achieve an ideal trade-off, given the rapid changes and extended range of values taken by $V_x$ in our problems, we introduce a symmetric logarithmic function in the gradient-related loss:
\begin{equation}
\minimize_{\theta^V} \quad 
\frac{1}{N_B}\sum\limits_{i=1}^{N_B} \left(\bar{V}_i - V(\Tilde x_i|\theta^V) \right)^2 +
k_S \left(\text{logsym}(V_{x,i}) - \text{logsym}(S_{x} V_{\Tilde x}(\Tilde x_i|\theta^V)) \right)^2,
\end{equation}
where the function logsym(x) is defined as
\begin{equation}
\text{logsym}(x) = 
    \begin{cases} 
      \log(x + 1) & \text{if } x \geq 0 \\
      -\log(-x + 1) & \text{if } x < 0
    \end{cases}
\end{equation}
The logsym function reduces the relative importance of potential large (either positive or negative) values in the gradient, which may arise when the network tries to approximate a discontinuous Value function (see the companion video for a visual illustration of this phenomenon).
\subsubsection{Differentiability of the Critic Network}
\label{ssec:critic_differentiability}
The original CACTO algorithm used ReLU activation functions for both actor and critic.
In CACTO-SL, we have observed that ReLU functions do not work well with Sobolev Learning. 
Indeed, ReLU functions lead to piecewise-linear approximations, for which the gradient may not represent well the local behavior of the function.
For this reason, in CACTO-SL the critic network uses SIREN (Sinusoidal Representation Networks) layers \cite{SIREN}. Characterized by its smooth and continuously differentiable nature, SIREN layers bring a crucial advantage to learning the gradient, preventing the generation of ill-behaved gradients from the loss term on the derivative mismatch. 

\subsubsection{Sample Efficiency}
\label{ssec:sample_efficiency}
By incorporating the Value gradient, the amount of information provided by each transition is much higher than in CACTO: each transition contains both the partial cost-to-go (1 scalar) and the gradient of the Value function (an $n$-dimensional vector). 
Therefore, while ensuring sufficient exploration of the state space is still necessary, we can expect fewer TO episodes to be needed for learning a good approximation of the Value function. This was empirically confirmed by the fact that we could benefit from increasing the ratio between the number of network updates and the number of TO episodes. 
In particular, with the aim of reducing the total number of TO episodes, we increased both the number of TO episodes before each update phase, $e_{update}$, and the number of updates performed at each algorithm iteration, $K$.
We start with a small value of $K$, because it is not useful to accurately learn the initial policy, which represents the TO solver. As the algorithm progresses, we increase $K$, performing more network updates after each TO-phase, as the accumulated data are expected to represent the Value function of a policy closer and closer to global optimality. 

This strategic adjustment enables a significant reduction in the number of TO episodes (by a factor of 3 to 10) and, as a consequence, in the computation time.

\subsubsection{Software for Trajectory Optimization and Multi-Body Dynamics}
\label{ssec:software}
While CACTO relied on \emph{Pyomo}~\cite{pyomo} for solving TO problems, CACTO-SL switched to \emph{CasADi}~\cite{casadi}, an open-source Automatic Differentiation framework for numerical optimization. 
By leveraging its symbolic framework and automatic differentiation capabilities, CasADi enables efficient computation of the cost function's derivatives, required for implementing Sobolev Learning. 
Moreover, \emph{CasADi} is compatible with \emph{Pinocchio}~\cite{Pinocchio}, a versatile rigid-body dynamics library, which freed us from the burden of hand-coding the system dynamics. 
Each TO problem is transcribed using collocation, and then solved with the nonlinear optimization solver \emph{IpOpt}~\cite{ipopt}.
Finally, to speed up the code, we parallelized the generation of the warm-start trajectories, the TO problems, and the computation of the cost-to-go and its gradient.

\section{Results}
\subsection{1-Dimensional Toy Problem}
\begin{figure}[tbp]
    \begin{minipage}[t]{0.45\textwidth}
        \centering
        \includegraphics[width=\linewidth]{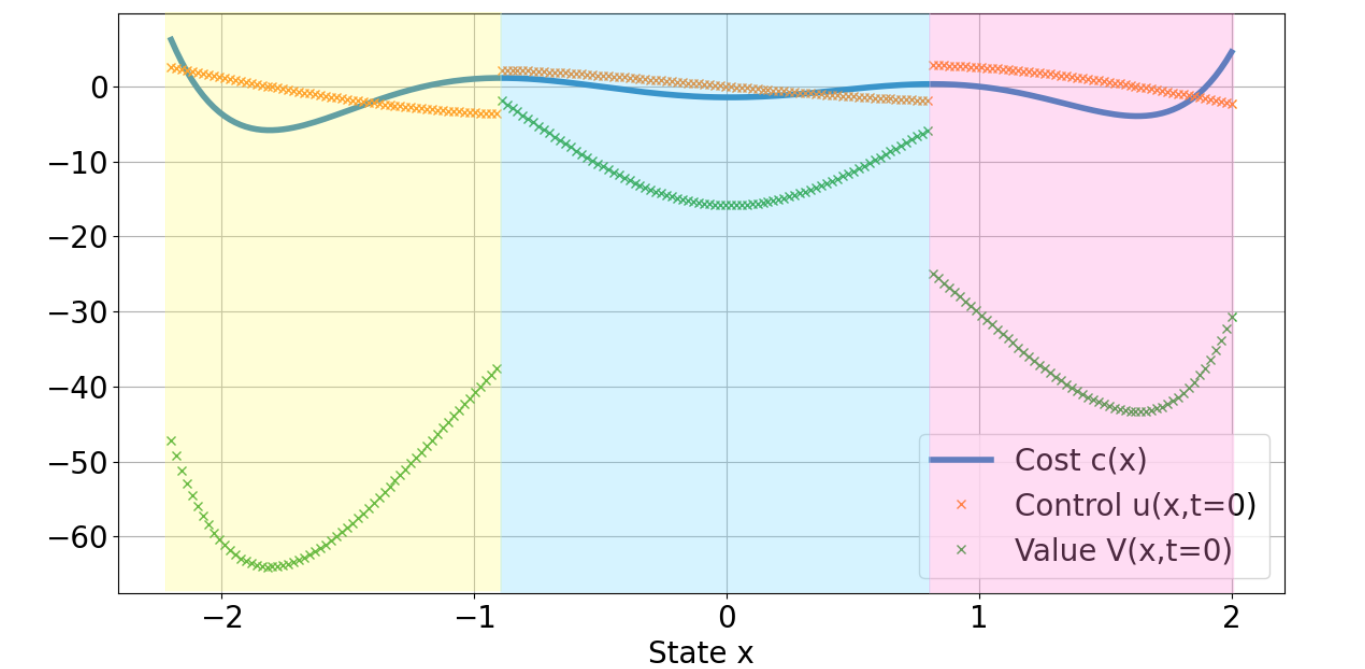}
        \caption{\vspace{-0.4cm}Cost, control and Value  obtained using a naive initial guess.}
        \label{fig:f1}
    \end{minipage}
    \hfill
    \begin{minipage}[t]{0.425\textwidth}
        \centering
        \includegraphics[width=\linewidth]{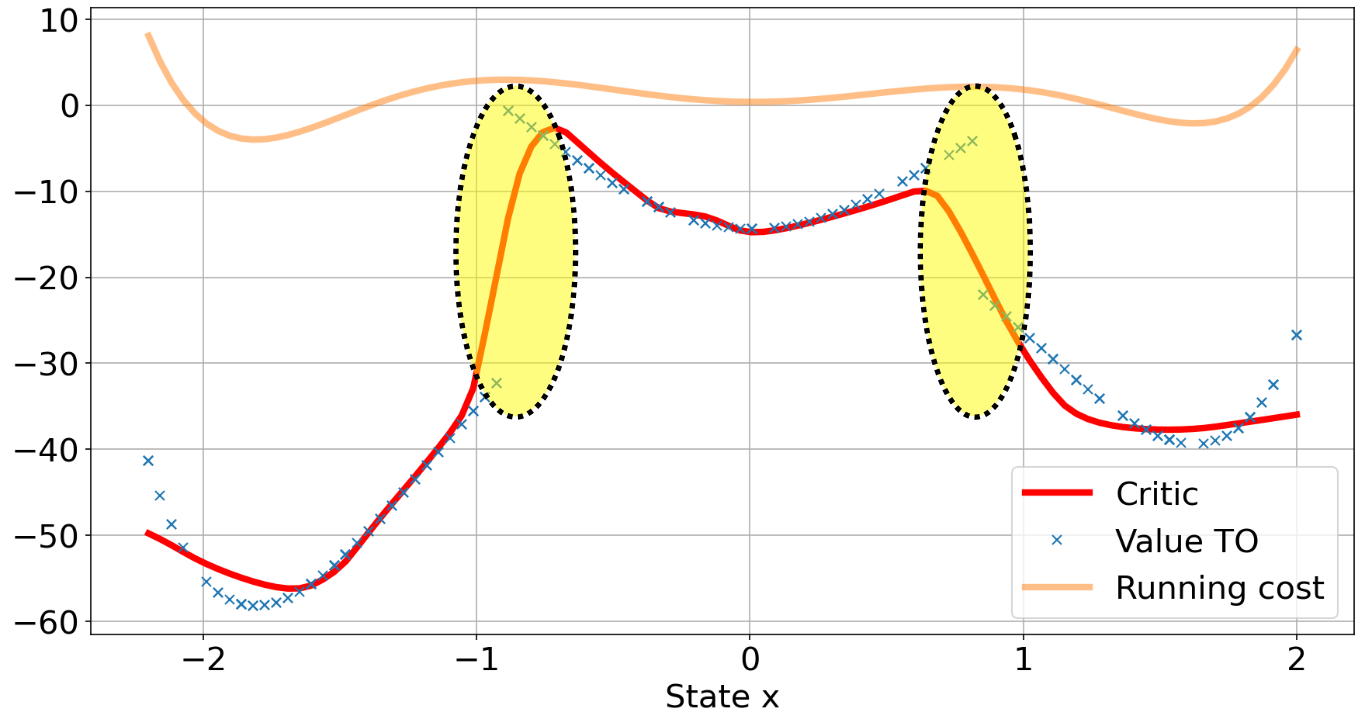}
        \caption{\vspace{-0.4cm}Critic smooths Value's discontinuities.}
        \label{fig:C1}
    \end{minipage}
\end{figure}

\begin{figure}[tbp]
    \begin{minipage}[t]{0.45\textwidth}
        \centering
        \includegraphics[width=\linewidth]{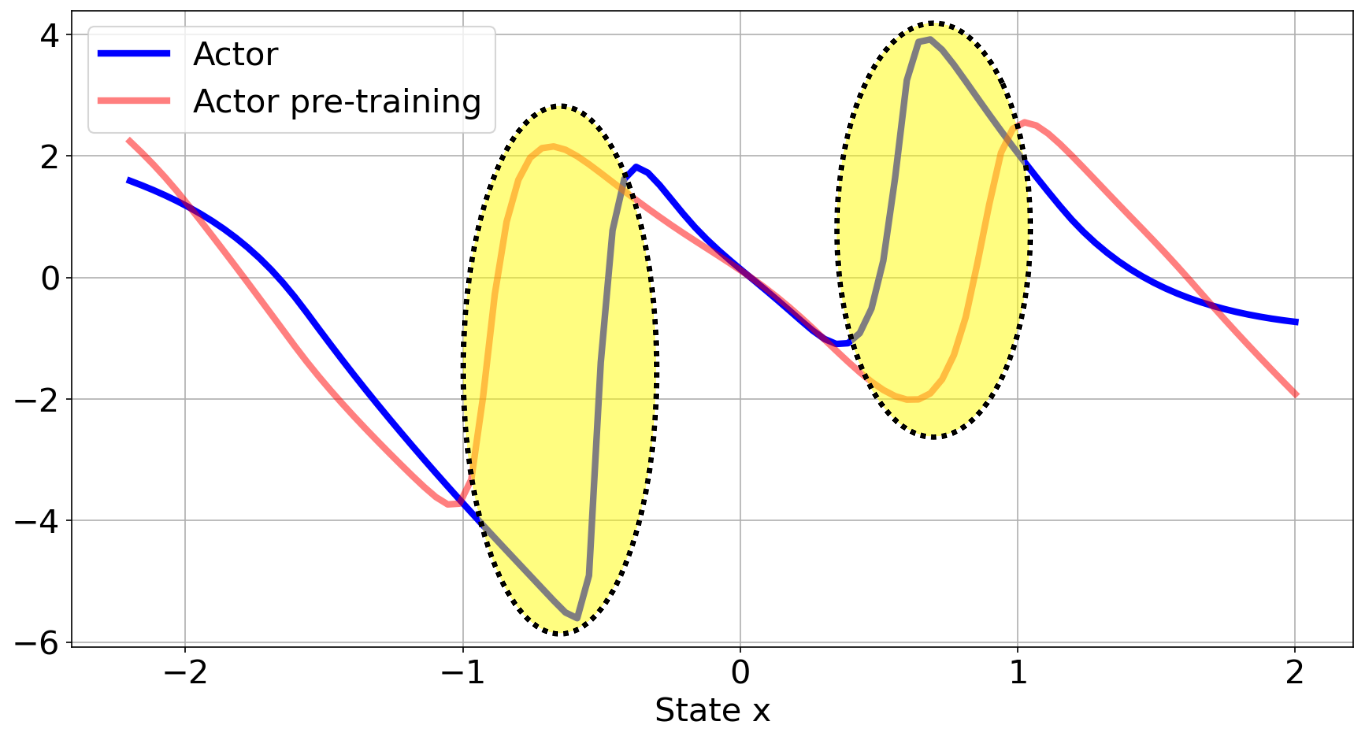}
        \caption{\vspace{-0.4cm}Actor after first iteration.}
        \label{fig:A1}
    \end{minipage}
    \hfill
    \begin{minipage}[t]{0.45\textwidth}
        \centering
        \includegraphics[width=\linewidth]{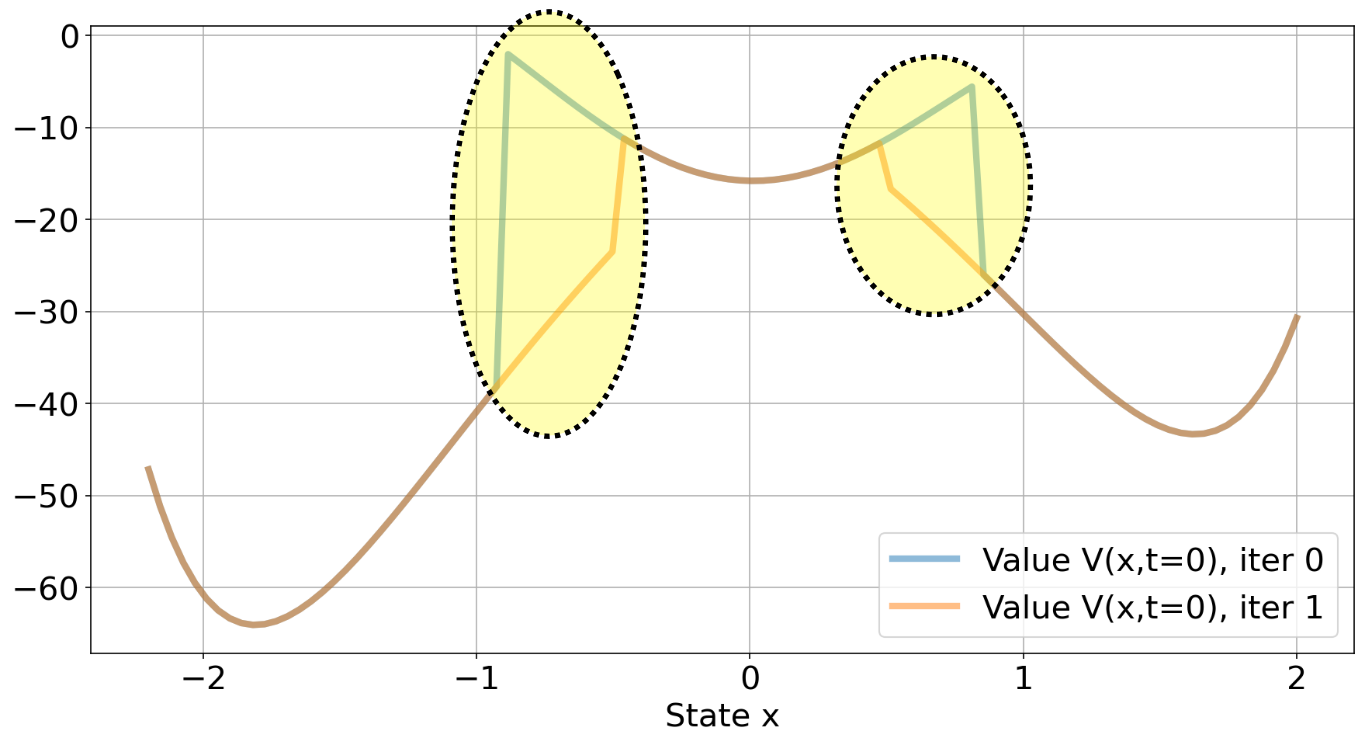}
        \caption{\vspace{-0.4cm}Value after first iteration.}
        \label{fig:V1}
    \end{minipage}
\end{figure}

This section exemplifies the behaviour of CACTO and the effect of Sobolev Learning using a 1D toy problem, with a 1D state, a single integrator dynamics and a cost function with three local minima (Fig.~\ref{fig:f1}).
Solving TO problems with a naive initial guess, highlights the presence of three basins of attraction (see Fig.~\ref{fig:f1}). Notice that $V(x)$ is not continuous at the boundaries of these basins. 
We now analyse the first iteration of CACTO. First, we solve 100 TO problems and train the critic network. Fig.~\ref{fig:C1} shows that the network smooths out the discontinuities because the critic network cannot represent discontinuous functions. 
After pre-training the actor network to imitate the TO control inputs, the actor network is trained to minimize the Q function. Fig.~\ref{fig:A1} and Fig.~\ref{fig:V1} show that the resulting actor provides a better policy, as the basin of attraction of the worst minimum is shrunk and the Value obtained by initializing TO with the trained actor policy has improved.

\begin{figure}[tbp]
    \begin{minipage}[t]{0.45\textwidth}
        \centering
        \includegraphics[width=\linewidth]{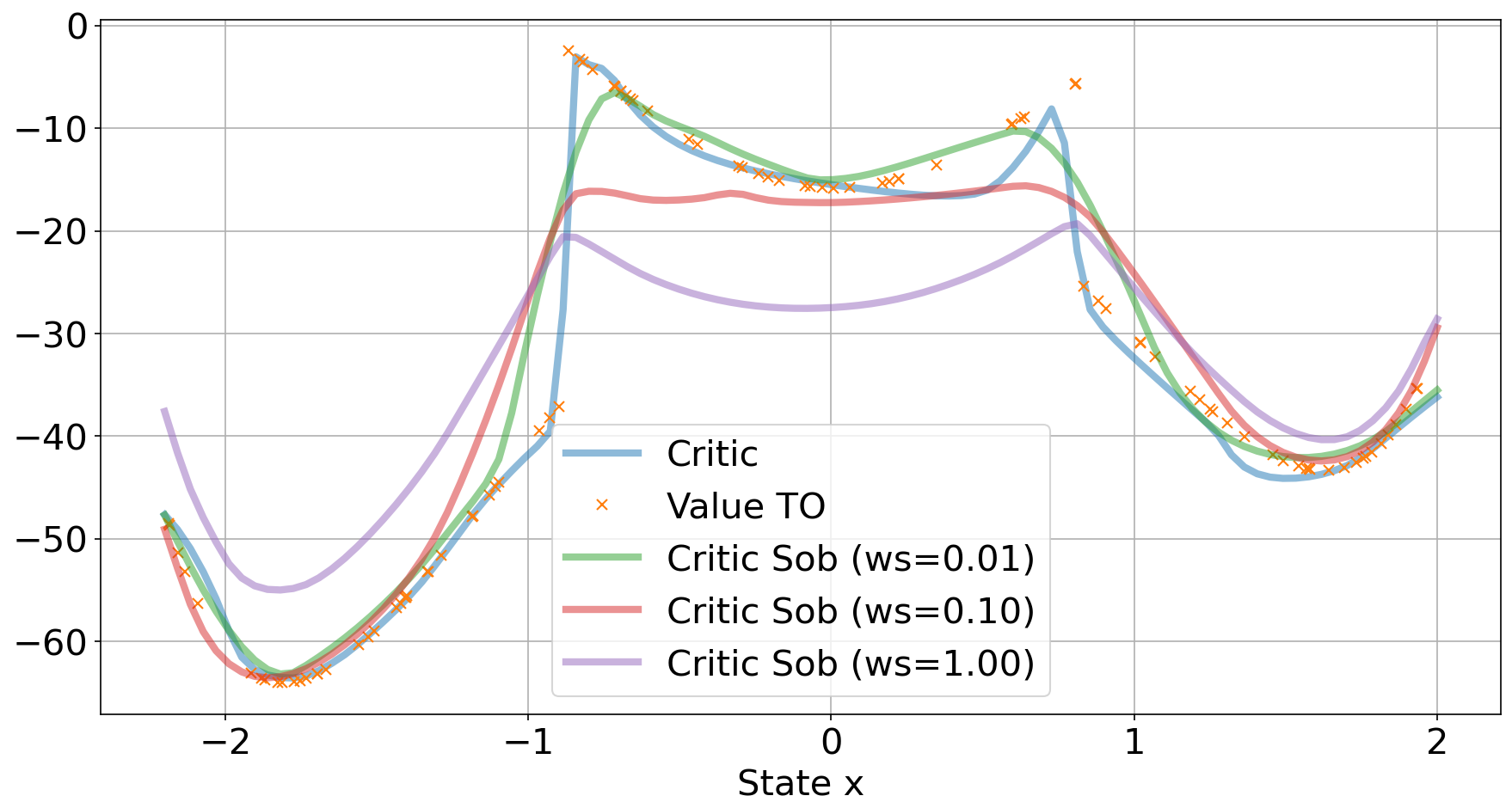}
        \caption{\vspace{-0.4cm}Effect of $k_s$ on critic learning: $V$.}
        \label{fig:SV}
    \end{minipage}
    \hfill
    \begin{minipage}[t]{0.45\textwidth}
        \centering
        \includegraphics[width=\linewidth]{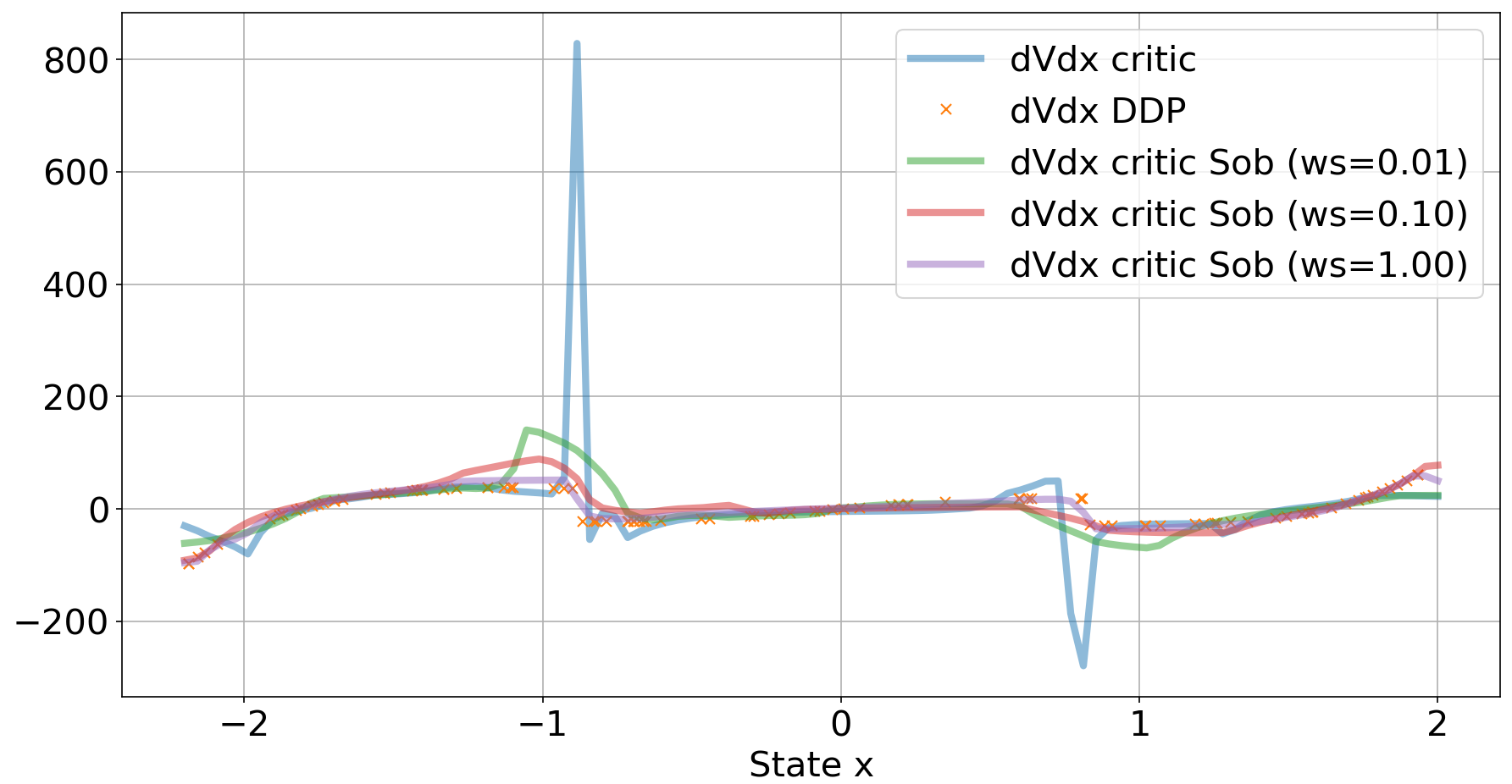}
        \caption{\vspace{-0.4cm}Effect of $k_s$ on critic learning: $V_x$.}
        \label{fig:SG}
    \end{minipage}
\end{figure}

We now analyse the effect of Sobolev Learning. As mentioned above, the critic is trained to predict the Value resulting from TO, which is typically discontinuous at the boundaries of the basins of attraction associated to the different local minima. This leads to a poor fit of the gradient in proximity of the discontinuities, and at the boundaries of the explored space. Introducing Sobolev Learning helps to smooth out the discontinuities because the critic tries to match also $V_x$, see Fig. \ref{fig:SV} and Fig. \ref{fig:SG}. 
As shown in Fig.~\ref{fig:SV} and Fig.~\ref{fig:SG}, increasing $k_S$, the smoothing  increases, which can help the actor to improve faster. However, $k_S$ should not be increased too much because it may lead to a complete removal of the discontinuities. This alters the real Value of the states and may prevent convergence to a globally optimal policy.


Interested readers can find a more detailed explanation in the \href{https://youtu.be/zAv8dP8itpM}{companion video}.

\subsection{Empirical Results}
This section presents our empirical results, with the goal to understand whether the use of Sobolev-Learning in CACTO is actually beneficial. The ability of CACTO-SL to provide good warm-starts to TO problems is analysed in the same scenarios presented in~\cite{CACTO}. The source code is available on the \href{https://github.com/gianluigigrandesso/cacto}{GitHub page of the project}. 

\begin{figure}[tbp]
     \makebox[\columnwidth][c]{\includegraphics[width = 0.5\columnwidth]
     {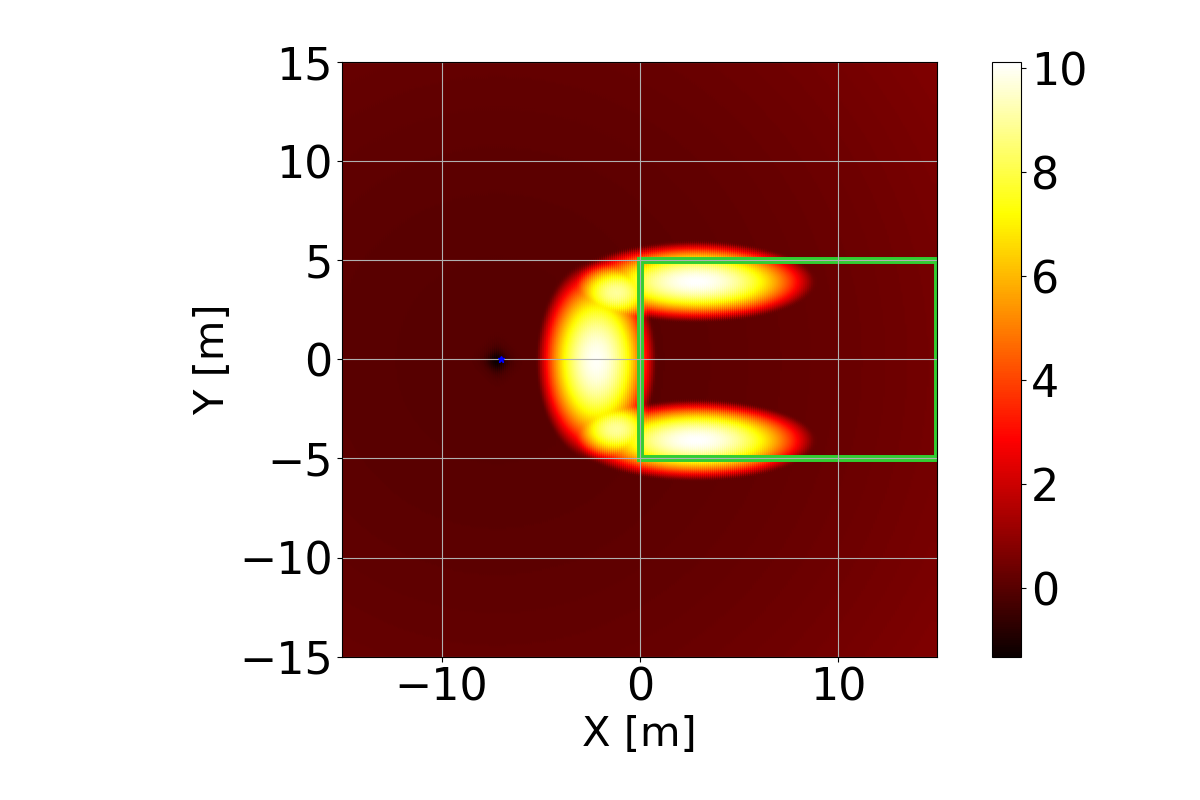}} 
    \caption{\vspace{-0.4cm}Cost function~\eqref{eq:running_cost} without the control effort term, considering a target at $p_g=[-7,0]$ with weights $w_d=10^{-3}$, $w_p=5$ and $w_{ob}=10^1$. The green rectangle delimits the \emph{Hard Region}.}
    \label{fig:CostFunction_comp}
\end{figure}

The analysed task consists in reaching in the shortest time a desired final position, while avoiding three ellipse-shaped obstacles, and minimizing the control effort. 
The task is described by the following running cost:
\begin{equation}
\begin{aligned}\label{eq:running_cost}
    l(x, u) =& \underbrace{w_{d}||p_{ee}-p_{g}||^2 \vphantom{\frac{w_{p}}{\alpha_1}}}_{l_1(x)} + \underbrace{-\frac{w_{p}}{\alpha_1}\ln(e^{-\alpha_1\left(\sqrt{(x_{ee}-x_{g})^2 + c_2}+\sqrt{(y_{ee}-y_{g})^2 + c_3} + c_4\right)}+1)}_{l_2(x)} + \\ & + \underbrace{\frac{w_{ob}}{\alpha_2}\sum_{i=1}^{3}\ln{(e^{-\alpha_2\left(\frac{(x_{ee}-x_{ob,i})^2}{(a_i/2)^2} + \frac{(y_{ee}-y_{ob,i})^2}{(b_i/2)^2} - 1\right)} + 1)}}_{l_3(x)} + \underbrace{w_{u}||u||_2^2 \vphantom{\sum_{i=1}^{3}}}_{l_4(u)} + \underbrace{\left\|\frac{u}{u_{max}}\right\|_2^{10} \vphantom{\sum_{i=1}^{3}}}_{l_5(u)}
\end{aligned}
\end{equation}
where $l_1$ penalizes the distance between $p_{ee} = (x_{ee}, y_{ee})$ (the x-y coordinate of the system's end-effector) and $p_g$ (the goal position to be reached); $l_2$ encodes a cost valley in the neighborhood of the goal position; $l_3$ penalizes collision with the three obstacles centered in $p_{ob,i}=(x_{ob,i},y_{ob,i})$ with axes $a_i$, and $b_i$;
the $w_*$'s are user-defined weights; $c_2$, $c_3$, $c_4$, $\alpha_1$, and $\alpha_2$ are the parameters of the softmax functions; $l_4$ and $l_5$ are the control regularization and penalty to ensure that the optimal controls are in the desired range. The terminal cost is equal to the running cost, except for $l_4$ and $l_5$.
Fig.~\ref{fig:CostFunction_comp} depicts the cost function, neglecting the control-effort term and control-penalty.

The task is designed to ensure the presence of many local minima. In particular, if the system starts from the \emph{Hard Region}, highlighted by a green rectangle in Fig.~\ref{fig:CostFunction_comp}, it is very hard for the solver to provide a globally optimal solution.
In all our tests, we compute the critic targets with TD(50) (i.e. $L=50$), which empirically led to a stable training of critic and actor.

\begin{figure}[tbp]
 {%
     \input{L4DC_figures/tikz_python/relu}
     \input{L4DC_figures/tikz_python/elu}
     \hfill
     \input{L4DC_figures/tikz_python/sine}
 }
 \caption{\vspace{-0.4cm}Comparison of critic networks based on ReLU (top), ELU (middle) and sinusoidal (bottom) activation functions at 1000, 5000 and 10000 updates for a point mass with single integrator dynamics.}
 \label{fig:compCN}
\end{figure}

\subsection{Comparison between activation functions}
We compared the critic network based on SIREN layers with the critic networks based on ReLU and based on ELU, another smooth activation layer, in combination with Sobolev Learning using $k_S=10^{3}$. Fig.~\ref{fig:compCN} shows that using the SIREN layers (or also ELU layers, but to a slightly lower extent) leads to a critic that clearly captures the lower values of states inside the C-shaped obstacle. On the other hand, this is completely overlooked by the critic using ReLU functions, even after 10000 iterations.

\subsection{Comparison between CACTO-SL and CACTO}
The comparison between CACTO-SL and CACTO is conducted with a 2D double integrator, and a Dubins car, using $k_s=10^3$. We compared the algorithms by computing the mean cost of the trajectories obtained by initializing TO with the policy's rollout. We used as initial states a grid of points over the analysed XY space, with mesh size equal to 1 m. 
The analysis focuses on the \emph{Hard Region} ($ x \in [0, 15]$ m and $y \in [-5, 5]$ m). For an easier interpretation of the results, the initial velocity of the system is set to zero. In the Dubins car test, the initial heading is randomly chosen. 
For each test, we performed 5 runs using the same seeds for both CACTO and CACTO-SL to minimize the effect of the network initialization.

\subsubsection{Double-integrator system}
We consider a 2D double integrator, with state vector $[x,y,v_x,v_y,t]\in \mathbb{R}^5$ and control vector $[a_x,a_y]\in \mathbb{R}^2$. 
The maximum episode length is 10 s. The target point is $p_g = [-7,0]$. 
In contrast to the previous version of CACTO, where we alternated $e_{update}=25$ TO episodes and $K=80$ network updates, in CACTO-SL we perform $e_{update}=200$ TO episodes and an increasing number of network updates, starting from $K=1000$ up to $K=15000$. 
Fig.~\ref{SR_HR_v0} reports the median (across 5 runs) of the mean cost (across initial conditions) starting from the \emph{Hard Region} against the number of TO episodes obtained initializing the TO problems with CACTO and CACTO-SL. 
In both cases, we stopped the training after 50k updates, when both algorithms have converged to a stable TO solution.
\begin{figure}[tb]
     \makebox[\columnwidth][c]
     {
\begin{tikzpicture}

\definecolor{darkgray176}{RGB}{176,176,176}
\definecolor{lightgray204}{RGB}{204,204,204}
\definecolor{mediumturquoise64181195}{RGB}{64,181,195}

\begin{axis}[
width=15 cm,
height=4.5 cm,
legend cell align={left},
legend style={fill opacity=0.8, draw opacity=1, text opacity=1, draw=lightgray204},
tick align=outside,
tick pos=left,
x grid style={darkgray176},
xlabel={$\#$ TO episodes $[10^4]$},
xmajorgrids,
xmin=0, xmax=10000,
xtick style={color=black},
xtick={0,2000,4000,6000,8000,10000},
xticklabels={0.0,0.2,0.4,0.6,0.8,1.0},%
scaled x ticks = {base 10:0},
y grid style={darkgray176},
ylabel={Mean cost},
ymajorgrids,
ymin=-47.4761538761168, ymax=36.4212402278878,
ytick style={color=black},
ytick={-60,-40,-20,0,20,40},
yticklabels={\ensuremath{-}6,\ensuremath{-}4,\ensuremath{-}2,0,2,4}
]
\path [draw=red, fill=red, opacity=0.2]
(axis cs:0,9.42711082140595)
--(axis cs:0,9.42711082140595)
--(axis cs:1500,-8.58932550762675)
--(axis cs:3000,-24.9714390265114)
--(axis cs:4500,-39.4775670682837)
--(axis cs:6000,-37.6227144380154)
--(axis cs:7500,-39.9187362881663)
--(axis cs:9000,-43.6069133904638)
--(axis cs:9000,-36.5735375305822)
--(axis cs:9000,-36.5735375305822)
--(axis cs:7500,-32.1497575067671)
--(axis cs:6000,-36.0837466065651)
--(axis cs:4500,-23.3163306377348)
--(axis cs:3000,3.64659746765195)
--(axis cs:1500,32.6077223140694)
--(axis cs:0,9.42711082140595)
--cycle;

\path [draw=mediumturquoise64181195, fill=mediumturquoise64181195, opacity=0.2]
(axis cs:0,9.42711082140595)
--(axis cs:0,9.42711082140595)
--(axis cs:400,-30.6398475389518)
--(axis cs:600,-38.3870572703966)
--(axis cs:800,-38.1067440234361)
--(axis cs:800,-39.343681240459)
--(axis cs:1000,-43.6626359622984)
--(axis cs:1000,-42.0033268099325)
--(axis cs:1000,-40.0198098006524)
--(axis cs:1000,-40.0198098006524)
--(axis cs:1000,-41.4209791741542)
--(axis cs:800,-35.3154565577383)
--(axis cs:800,-31.203261012663)
--(axis cs:600,-30.6957660540385)
--(axis cs:400,6.41207801976498)
--(axis cs:0,9.42711082140595)
--cycle;

\addplot [semithick, red, mark=*, mark size=2, mark options={solid}]
table {%
0 9.42711082140595
1500 -3.10181906264594
3000 -21.6575007606401
4500 -37.2878873018779
6000 -36.4125796757742
7500 -37.5947066624804
9000 -41.6320159396597
};
\addlegendentry{CACTO}
\addplot [semithick, mediumturquoise64181195, mark=*, mark size=2, mark options={solid}]
table {%
0 9.42711082140595
400 -28.6771863650541
600 -35.8393133183918
800 -37.6525680834291
800 -37.1582202684398
1000 -42.2536494160335
1000 -41.1526044597217
};
\addlegendentry{CACTO-SL}
\end{axis}

\begin{axis}[
at={(0cm,-4.75cm)},
width=15 cm,
height=4.5 cm,
legend cell align={left},
legend style={
  fill opacity=0.8,
  draw opacity=1,
  text opacity=1,
  at={(0.97,0.03)},
  anchor=south east,
  draw=lightgray204
},
tick align=outside,
tick pos=left,
x grid style={darkgray176},
xlabel={$\#$ TO episodes $[10^4]$},
xmajorgrids,
xmin=0, xmax=25000,
xtick style={color=black},
xtick={0,5000,10000,15000,20000,25000},
xticklabels={0.0,0.5,1.0,1.5,2.0,2.5},
scaled x ticks = {base 10:0},
y grid style={darkgray176},
ylabel={Mean cost},
ymajorgrids,
ymin=-7.25931853440151, ymax=25.4216494705021,
ytick style={color=black},
ytick={-10,0,10,20,30},
yticklabels={\ensuremath{-}1,0,1,\phantom{-}2,3}
]
\path [draw=red, fill=red, opacity=0.2]
(axis cs:0,14.1906216316322)
--(axis cs:0,14.1906216316322)
--(axis cs:3125,14.9413425498474)
--(axis cs:6250,16.3916816908464)
--(axis cs:9375,-0.119218225596091)
--(axis cs:12500,6.28161251577283)
--(axis cs:15625,8.75401490731845)
--(axis cs:18750,6.94825320394344)
--(axis cs:21875,11.8274936975223)
--(axis cs:25000,7.14263626504745)
--(axis cs:25000,23.9361509248247)
--(axis cs:25000,23.9361509248247)
--(axis cs:21875,20.6217355361493)
--(axis cs:18750,16.1153860223464)
--(axis cs:15625,15.8184985422649)
--(axis cs:12500,14.7304240176826)
--(axis cs:9375,9.42929584675406)
--(axis cs:6250,17.2555926667374)
--(axis cs:3125,19.7213911674942)
--(axis cs:0,14.1906216316322)
--cycle;

\path [draw=mediumturquoise64181195, fill=mediumturquoise64181195, opacity=0.2]
(axis cs:0,14.1906216316322)
--(axis cs:0,14.1906216316322)
--(axis cs:2000,2.98143603100036)
--(axis cs:3000,2.87531013613734)
--(axis cs:3500,-2.25081479446623)
--(axis cs:4000,-5.77381998872407)
--(axis cs:5000,1.09973549562849)
--(axis cs:5500,-2.08918070696696)
--(axis cs:6000,-1.28764775416439)
--(axis cs:7000,-2.93610148206266)
--(axis cs:7000,3.19557992896053)
--(axis cs:7000,3.19557992896053)
--(axis cs:6000,6.93473144377871)
--(axis cs:5500,3.57386124674267)
--(axis cs:5000,2.43625978146774)
--(axis cs:4000,-1.25043846077493)
--(axis cs:3500,6.47531133342001)
--(axis cs:3000,8.74481807354454)
--(axis cs:2000,10.394019489187)
--(axis cs:0,14.1906216316322)
--cycle;

\addplot [semithick, red, mark=*, mark size=2, mark options={solid}]
table {%
0 14.1906216316322
3125 16.9838501923382
6250 16.6896048374361
9375 5.80871827461598
12500 12.2126161420586
15625 12.8052983718282
18750 14.2162378498351
21875 16.7401664451065
25000 17.467873385915
};
\addlegendentry{CACTO}
\addplot [semithick, mediumturquoise64181195, mark=*, mark size=2, mark options={solid}]
table {%
0 14.1906216316322
2000 5.75999623107869
3000 7.73058220809977
3500 2.4093503670753
4000 -2.83541114820285
5000 1.55712803705467
5500 -0.0909720516137063
6000 0.674663216174094
7000 0.219401110888924
};
\addlegendentry{CACTO-SL}
\end{axis}

\end{tikzpicture}}
    \caption{\vspace{-0.4cm}Median (across 5 runs) of the mean cost (across initial conditions) starting from the \emph{Hard Region} with zero initial velocity against the number of TO episodes for a point mass with double integrator dynamics (top) and a Dubins car (bottom). The shaded area represents the area between the first and third quartiles.}
    \label{SR_HR_v0}
\end{figure}

Fig.~\ref{SR_HR_v0} shows that CACTO-SL has led to a 92$\%$ reduction in the number of TO problems. As regards the computation time, it is important to recall that it heavily depends on the algorithm implementation, on the hardware, and on the number of cores used in the parallelization. Parallelizing the execution on 10 cores, the decrease of TO problems has resulted in a reduction of just 7$\%$ of the total execution time because the system is quite simple. However, without parallelization, the total computation time is reduced from 111 to 49 minutes. Moreover, Fig. \ref{SR_HR_v0} shows a significant reduction in the variance across the five runs. This outcome can be attributed to the enhanced training of the critic through Sobolev Learning. The improved training results in a more consistent set of critic networks. As a consequence, the actors converge towards similar policies, leading to a more uniform initial guess across runs.

\subsubsection{Car system}
The second problem involves a Dubins car modelled as a point mass. The 6D state vector consists in the displacement along x and y, angular displacement, linear velocity, linear acceleration and time. The 2D control vector contains the angular velocity and the linear jerk. The car should reach the target point $p_g = [-7,0]$ m. No penalty is added on its orientation. We updated the networks every $e_{episode}=500$ TO episodes, using an increasing number of updates, starting from $K=1000$ and reaching $K=15000$. In CACTO instead the ratio between TO episodes and network updates was 25/160. Fig.~\ref{SR_HR_v0} shows the results. 
In this case, 170k updates have been performed.

In this case, CACTO-SL leads to a notable reduction in the number of TO problems solved and, consequently, in the computation time, from 10 h to 6.30 h (parallelizing the TO problems on 10 cores). Note that the reduction in computation time increases significantly also with parallelization, even though the Dubins car has only one more state than the 2D double integrator. Also in this test, Sobolev Learning plays an important role in achieving smaller variance across the runs.

\section{Conclusions}
This paper presented CACTO-SL, an extension of the CACTO algorithm, featuring Sobolev Learning. 
CACTO-SL efficiently computes the gradient of the Value function using the backward pass of the DDP algorithm, and exploits this information to improve the training of the critic network. 
Our results show that this makes CACTO-SL significantly more sample efficient than CACTO, reducing the number of TO episodes by a factor ranging from 3 to 10. This also leads to almost halving the computation time required to reach a stable solution. Moreover, our results show that CACTO-SL reduces the variance of the algorithm across different runs, and helps TO to find better minima.

In the future, we plan to switch to a BoxDDP backward pass~\cite{Tassa2014}, which would allow us to consider hard bounds on the control inputs.
Moreover, we are exploring different approaches to bias the sampling of initial states towards more informative state regions, which should improve even further the sample efficiency of CACTO and allow us to scale better to more complex systems.


\acks{This work was supported by PRIN project DOCEAT (CUP n. E63C22000410001) and by the European Union - Next Generation EU, Mission 4 Component 2 - STARLIT (CUP n. E53D23001130006).}

\bibliography{references.bib}

\end{document}